\documentclass[11pt,letterpaper]{article}

\usepackage[margin=1in]{geometry}
\usepackage{times}
\usepackage{microtype}
\usepackage{setspace}
\usepackage{natbib}
\usepackage{amsmath,amssymb}
\usepackage{graphicx}
\usepackage{float}
\usepackage[section]{placeins}
\usepackage{hyperref}
\usepackage{xcolor}
\usepackage{parskip}
\usepackage{tcolorbox}
\usepackage{booktabs}

\hypersetup{colorlinks=true, linkcolor=blue!60!black, citecolor=blue!60!black, urlcolor=blue!60!black}
\newcommand{\doi}[1]{\href{https://doi.org/#1}{doi:\ #1}}
\tcbset{
    boxstyle/.style={
        colback=gray!5, colframe=gray!50!black, fonttitle=\bfseries\small,
        boxrule=0.5pt, arc=2pt, left=6pt, right=6pt, top=4pt, bottom=4pt,
    }
}

\title{\Large\bfseries Minimal Oversight: Uncertainty-Aware Governance for Delegated AI Systems}
\author{Carlos R.\,B.\ Azevedo\\[4pt]
\small\textit{Independent Researcher, S\~ao Paulo, Brazil}\\[2pt]
\small\texttt{renatoaz@gmail.com}}
\date{}

\begin{document}
\raggedbottom
\widowpenalty=10000
\clubpenalty=10000
\maketitle


\begin{abstract}
\noindent AI systems increasingly delegate decisions to specialized models,
evaluators, tools, and supervisory controllers. The central AI problem is no
longer only model accuracy, but uncertainty-aware governance: how much
autonomy to grant, which evidence should calibrate trust, what performance
ceiling a delegated AI system can sustain, and when human intervention becomes
necessary. We propose the Minimum Sufficient Oversight Principle (MSO), a variational
principle for principled autonomy delegation: minimize governance burden on
the Fisher information manifold subject to a delivery constraint. The resulting
Euler--Lagrange solution yields a water-filling allocation of governed
delegation across the task space. Building on a revealed-action governed
delegation channel model, we prove a capacity theorem for stationary
symbolwise review policies, derive a local first-order approximation relating
workflow complexity to quality degradation, and give a drift-dominated
autonomy-time scaling law linking intervention timing to effective capacity,
complexity, and drift. Within this framework, masking appears as a structural
AI-governance pathology: corrected performance can hide the competence signal
needed to calibrate trust. Synthetic simulations and a semi-real reconstructed
workflow support design prescriptions including upstream-first correction,
sensitivity-based intervention, and explicit feasibility checks before
autonomy is expanded. The result is a computable framework for uncertainty,
planning, and oversight in delegated AI systems. A companion Python package is
available at \url{https://github.com/crbazevedo/delegation-lab}.
\end{abstract}


\bigskip

\noindent\small\textbf{Working symbols used throughout.}\;
\textbf{State:} $\sigma_{\mathrm{raw}}$ (raw competence),
$\sigma_{\mathrm{corr}}$ (delivered quality), $H(W)$ (workflow complexity).\;
\textbf{Control:} $\alpha(x)$ (governed-delegation allocation), $d(x)$ (delegated-scope indicator),
$K$ and $K/N$ (discrete review capacity and coverage), $B$ (average review budget).\;
\textbf{Limits:} $C_{\mathrm{op}}$ (operational quality ceiling),
$C_{\mathrm{del}}(B)$ (governed-channel capacity),
$B_{\mathrm{eff}}$ (effective autonomy buffer),
$T^{*}_{\mathrm{auto}}$ (autonomy time).\;
\textbf{Diagnostics:} $M^{*}$ (masking index),
$\mathrm{DC}(v)$ (delegation centrality).
\normalsize

\medskip

\noindent\textbf{Notation discipline.}\; We use $K$ for \emph{discrete}
corrector capacity in synchronous or queueing-style models, so $K/N$
is the fraction of items that can be reviewed in a cycle. We use $B$
for the \emph{average review-cost budget} in the governed-channel
model of Definition~1. When a governance policy is written as
$\pi=(K,\rho,\phi)$, its $K$ component denotes the same discrete
review-capacity resource as in $K/N$, specialized to the policy level.
We reserve $C_{\mathrm{op}}$ for the operational quality ceiling used
in design and experiments, and $C_{\mathrm{del}}(B)$ for the
information-theoretic capacity of the governed delegation channel.

\bigskip


\noindent Modern AI systems increasingly rely on \emph{delegated decision
structures}: one model proposes an answer, another evaluates it, tools supply
evidence, and a supervising policy decides what may proceed. The central AI
design problem is no longer only how accurate an individual model is. It is
how uncertainty should be represented, how trust should be calibrated from
evidence, how much autonomy to grant, where to place oversight, and when
human intervention becomes necessary as the system evolves.

This paper develops a theory of \textbf{principled autonomy delegation} for such systems. The aim is to make delegation computable rather than heuristic. Given an agent's demonstrated competence, the difficulty structure of the task space, and a limited governance budget, the theory should tell us four things: how oversight should be allocated, what performance ceiling the delegated system can actually achieve, how workflow complexity degrades that ceiling, and how long the system can operate before intervention is required.

We formalize this with the \textbf{Minimum Sufficient Oversight Principle} (MSO):
minimize total governance burden on the Fisher information manifold subject to
a delivery constraint. The resulting allocation is a water-filling solution
over the task space. From this principle, and from an explicit channel model
of governed delegation, we derive a family of results: an optimal
governed-delegation allocation, a stationary delegation-capacity theorem, a
local process-complexity sensitivity law, and a drift-dominated autonomy-time
scaling law from first-passage theory. These quantities support a standard AI
workflow: represent uncertainty, plan feasible autonomy, monitor drift, and
intervene when the evidence no longer supports delegation. Graph topology
appears only because it determines how uncertainty, correction, and error
propagation affect trust calibration.

Within this broader framework, one structural failure mode appears immediately. In delegated systems, the process that preserves output quality can also destroy the information needed to calibrate trust. When corrected performance is used as the basis for authority, oversight can hide deterioration in the producing agent. We call this \textbf{masking}. It is not the main object of the theory; it is the first pathology the theory exposes and resolves by separating raw competence from corrected output quality.

The paper's thesis is therefore broader than measurement correction. It is that autonomy delegation can be governed by a principled variational and information-theoretic framework, rather than by ad hoc routing rules, intuition, or static safety margins. Masking is one consequence of that framework. The larger contribution is a calculable design logic for delegated systems: how topology, workflow complexity, review capacity, and measured competence jointly determine what can be delegated, under what oversight, for how long.

A concrete example helps fix ideas. Consider an AI-assisted
software-delivery workflow: a generator proposes code, an evaluator inspects
it, testing and security tools provide evidence, and a supervisory gate decides
what ships. The designer must decide which evidence calibrates trust, how much
autonomy each stage receives, whether the system can support the required
quality target at all, and how often humans must intervene. The MSO turns
those questions into explicit quantities rather than heuristics.

\medskip

\noindent\textbf{Category scope.}\; The paper is positioned as a contribution
to artificial intelligence, especially uncertainty in AI, expert-system
governance, and planning under limited evidence. It uses delegated workflows
and graph motifs as a modeling substrate, but it does not study coordination
protocols or distributed-agent negotiation as its primary object.

\medskip

\noindent\textbf{A note on theoretical status.}\; The results in this
paper span a range of rigor. We mark each as: \textbf{Theorem} (proved
under stated assumptions), \textbf{Proposition} (derived under
approximations), or \textbf{Empirical Law} (observed in simulations).
This demarcation appears at first use of each result and is summarized
in Section~4.

\medskip

\noindent\textbf{Outline.}\; Section~1 develops the theory: the MSO,
the Fisher metric, the Return Operator, delegation graphs, the
Euler--Lagrange water-filling solution, the delegation capacity,
process entropy, and the autonomy time. Section~2 connects to existing
frameworks. Section~3 presents numerical validation across eight
conditions. Section~4 discusses implementation, limitations, and conclusions.


\section{The Principle}

We begin with the operational setup: a delegation, its measurements,
and the principle that governs it. The theory rests on three primitives
(distinction, record, agency); here we proceed directly from their operational
consequences. The term distinction is used in the operational sense of drawing
a boundary between observable states, not as a dependence on Spencer-Brown's
calculus of indications \citep{spencerbrown1969}.

\subsection*{The delegation setup}

A \textbf{delegation} consists of a principal~$A$, an agent~$B$, a
scope~$S$ (the set of tasks $B$ is authorized to perform), and a
corrector~$C$ that observes $B$'s outputs and applies corrections.
At each point $x \in S$, the agent produces outcomes; the corrector
reviews a subset. Both raw and corrected outcomes enter the record
state $R(t)$. From these records we compute two quantities that will
drive the entire theory:

\begin{itemize}\setlength\itemsep{2pt}
    \item \textbf{Evidential support} $\sigma_{\mathrm{raw}}(x,t)$: the
    empirical success rate of $B$'s \emph{uncorrected} outcomes at~$x$,
    computed as the running fraction of correct outputs in the accumulated
    records $R(t)$. This is a Bernoulli parameter in $[0,1]$ that measures
    $B$'s demonstrated competence. (It is related to the mutual information
    by $I(R;\,\mathrm{outcome}) = 1 - H(\sigma_{\mathrm{raw}})$, where $H$
    is the binary entropy function.)
    \item \textbf{Corrected support} $\sigma_{\mathrm{corr}}(x,t)$: the
    empirical success rate computed from outcomes that include the
    corrector's interventions. This measures the system's output quality.
\end{itemize}

\noindent Together, these two quantities define the minimal informational
state needed for principled autonomy delegation: one signal for agent
competence and one for delivered system quality. Their separation also
resolves the masking failure mode introduced above. Authorization must
be based on $\sigma_{\mathrm{raw}}$---not on $\sigma_{\mathrm{corr}}$,
which conflates the agent's competence with the corrector's diligence.

\subsection*{The Minimum Sufficient Oversight Principle}

Why minimize oversight cost? The intuition is that oversight is
\emph{expensive}: every unit of review, monitoring, or corrective
capacity spent on one scope point is a unit unavailable elsewhere. This is an
informational and organizational cost, not a thermodynamic lower bound in
Landauer's sense \citep{landauer1961}. A
code reviewer spending time re-checking a reliable module has less
capacity for the fragile one. Minimizing oversight cost means spending
the review budget \emph{efficiently}---concentrating attention where it
produces the most quality improvement, not where the agent is already
competent. This is not a philosophical preference---it is the
variational structure of the problem, analogous to Shannon's
water-filling (power is allocated to channels proportionally to their
capacity for improvement, not uniformly).

\medskip

The central control variable is the \textbf{governed-delegation intensity}
$\alpha(x,t) \in [0,1]$. It is the fraction or weight of task class $x$
allowed to be handled by the delegated system during interval $t$ under the
available governance regime. Thus $\alpha$ is not the same object as the
discrete review fraction $K/N$ or the average review budget $B$: $K$ and $B$
describe the scarce review resource, while $\alpha$ describes how much
delegated workload is placed under that resource. A baseline trust policy may
map evidential support to a maximum allowed intensity,
\begin{equation}\label{eq:governance}
    \alpha(x,t) \leq \alpha_{\max}(x,t)
    = G\bigl(\sigma_{\mathrm{raw}}(x,t)\bigr),
\end{equation}
where $G$ is monotone non-decreasing. The MSO then optimizes the actual
allocation $\alpha$ beneath this ceiling. This distinction avoids conflating
two different questions: how much authority a task class is allowed to receive
and how scarce oversight effort should be distributed within that allowed
region.

\begin{tcolorbox}[colback=blue!3, colframe=blue!40!black,
    fonttitle=\bfseries\small, boxrule=0.5pt, arc=2pt,
    left=6pt, right=6pt, top=4pt, bottom=4pt,
    title=The Minimum Sufficient Oversight Principle (MSO)]
The delegation allocates the minimum sufficient governance burden over scope
and time, subject to a delivery constraint:
\begin{equation}\label{eq:mso}
    \min_{\alpha}\; \int_0^T \!\!\int_S \alpha(x,t)^{2}\,
    \sqrt{\det g(x,t)}
    \;\mathrm{d}x\, \mathrm{d}t
    \quad\text{subject to}\quad
    \int_S \alpha\,\sigma_{\mathrm{raw}}\;\mathrm{d}x
    \geq p_{\min}\,|S|.
\end{equation}

\emph{In words:} allocate only the oversight burden---measured in the Fisher
information geometry---sufficient to meet the quality target. The
quadratic cost $\alpha^{2}$ reflects diminishing returns: concentrating
all governed workload at one point is more expensive than spreading it, just as
concentrating all power on one channel is inefficient in Shannon's
water-filling.
\end{tcolorbox}

\noindent\emph{Why quadratic cost?}\; A linear cost $\int \alpha
\sqrt{g}\,\mathrm{d}x$ would yield a knapsack-like solution that
selects points by the ratio $\sqrt{g}/\sigma$ rather than allocating
proportionally. The quadratic form $\alpha^{2}$ models the realistic
property that the marginal cost of oversight is \emph{increasing}:
doubling review effort at a point costs four times as much. This is
the standard assumption in resource allocation theory and produces the
water-filling solution that is the hallmark of optimal allocation in
information theory \citep{cover2006}.

\medskip

\noindent\emph{Scope and geometry.}\; The MSO above treats the delegated
scope $S$ as fixed. If scope itself is a control variable, introduce
$d(x)\in\{0,1\}$ and optimize
\[
    \min_{\alpha,d}\int_S d(x)\alpha(x)^2\sqrt{\det g(x)}\,dx
    \quad \text{s.t.}\quad
    \int_S d(x)\alpha(x)\sigma(x)\,dx
    \ge p_{\min}\!\int_S d(x)\,dx .
\]
This endogenous-scope extension is useful, but it also exposes a
cherry-picking issue: without a coverage or task-value constraint such as
$\int_S d(x)w(x)\,dx\ge W_{\min}$, the optimum may delegate only a small
high-competence subset. The short version of the paper therefore focuses on
the fixed-scope problem and treats scope selection as an outer design
constraint.

The volume element $\sqrt{\det g}$ is the Fisher information metric
\citep{cencov1982,amari2000}. For Bernoulli outcomes with success
probability $\sigma$,
\begin{equation}\label{eq:fisher}
    g(\sigma)=\frac{1}{\sigma(1-\sigma)} .
\end{equation}
This is the metric in the local coordinate $\sigma$. If the task coordinate is
an external coordinate $x$ with success curve $\sigma(x)$, the pullback metric
is $g_x=(\partial_x\sigma)^2/[\sigma(1-\sigma)]$. The discrete models below
index each cell by its empirical competence value, so equation~(\ref{eq:fisher})
is the intended local metric.

\subsection*{The Return Operator}

The delegation evolves through a cyclic process we call the
\textbf{Return Operator}~$R$: $B$ operates $\to$ $C$ corrects $\to$
records accumulate $\to$ $\sigma$ updates $\to$ $G$ updates $\alpha$
$\to$ scope adjusts. We derive the dynamics of $\sigma$ from first
principles.

At each point $x$, in each time interval $\mathrm{d}t$, the agent
produces an outcome with probability $\eta(x,t)\,\mathrm{d}t$ of being
observed, where $\eta$ is the observation rate. Each observation updates
$\sigma$ toward the agent's true competence $\sigma_{\mathrm{skill}}(x)$,
because the Bayesian posterior on competence, given a new outcome, moves
in the direction of the data. Simultaneously, old records lose relevance
at rate $\delta$---the environment shifts, skills change, and stale
evidence no longer applies. When stale evidence is removed, the estimate
should not necessarily decay to zero; it should decay toward a prior
support level $\sigma_0(x)$ (for example, $1/2$ for an uninformative
Bernoulli prior, or $0$ for a conservative ``no demonstrated support''
score). Combining these two effects:
\begin{equation}\label{eq:sigma_dynamics}
    \frac{\partial\sigma_{\mathrm{raw}}}{\partial t}
    = \underbrace{\eta(x,t)\bigl[\sigma_{\mathrm{skill}}(x) -
    \sigma_{\mathrm{raw}}(x,t)\bigr]}_{\text{learning: $\sigma$ moves
    toward truth}}
    \;-\; \underbrace{\delta\bigl[\sigma_{\mathrm{raw}}(x,t)-\sigma_0(x)\bigr]}_{\text{forgetting:
    $\sigma$ relaxes to prior support}}.
\end{equation}
This is a linear relaxation equation, structurally identical to a
leaky integrator in neuroscience or a first-order low-pass filter in
engineering. The agent ``charges up'' toward $\sigma_{\mathrm{skill}}$
through observations and relaxes toward the prior through decay. The
balance determines the equilibrium.

\medskip

\noindent\textbf{Fixed point.}\; \emph{Assumptions: binary outcomes
(Bernoulli), stationary $\sigma_{\mathrm{skill}}$, constant $\eta$,
$\delta$, and $\sigma_0$.} Setting $\partial\sigma/\partial t = 0$ and solving:
\begin{equation}\label{eq:fixed_point}
    \sigma_{\mathrm{raw}}^{*}
    = \frac{\eta\;\sigma_{\mathrm{skill}} + \delta\,\sigma_0}{\eta + \delta}.
\end{equation}
The intuition is immediate: $\sigma^{*}$ is a weighted average of the
truth ($\sigma_{\mathrm{skill}}$) and the prior support $\sigma_0$, with
weights $\eta$ (observation) and $\delta$ (decay). When observations
are frequent relative to decay ($\eta \gg \delta$), $\sigma^* \to
\sigma_{\mathrm{skill}}$---the system learns the truth. When decay
dominates ($\delta \gg \eta$), $\sigma^* \to \sigma_0$---the system
returns to its prior. The numerical examples below use the conservative
support convention $\sigma_0=0$, under which equation~(\ref{eq:fixed_point})
reduces to the simpler expression
$\sigma_{\mathrm{raw}}^{*}=\eta\sigma_{\mathrm{skill}}/(\eta+\delta)$.
For $\eta = 10$, $\delta = 2$, and $\sigma_{\mathrm{skill}} = 0.80$,
this gives $\sigma^{*}=0.667$. With an uninformative Bernoulli prior
$\sigma_0=0.5$, the same parameters would give $\sigma^{*}=0.750$.

\medskip

\noindent\textbf{Convergence.}\; The linearized dynamics around the
fixed point decay at rate $\eta + \delta$---the spectral gap of $R$.
The distance to equilibrium decreases as
$|\sigma(t) - \sigma^{*}| \propto e^{-(\eta+\delta)t}$. Convergence
is faster when the observation rate is high and the evidence memory is
short. The time constant itself does not scale with the number of scope
points $N$; what improves with larger $N$ is the statistical uncertainty
of an average over independent cells, whose variance shrinks as $O(1/N)$.

\medskip

\noindent\textbf{The corrector's effect on $\sigma$.}\; The corrector $C$
catches a fraction $c$ of the agent's errors. The \emph{corrected}
fixed point is:
\begin{equation}\label{eq:sigma_corr}
    \sigma_{\mathrm{corr}}^{*}
    = \sigma_{\mathrm{raw}}^{*} + (1 - \sigma_{\mathrm{raw}}^{*}) \times c.
\end{equation}
The masking index follows directly:
$M^{*} = \sigma_{\mathrm{corr}}^{*} / \sigma_{\mathrm{raw}}^{*}
= 1 + (1 - \sigma_{\mathrm{raw}}^{*}) c / \sigma_{\mathrm{raw}}^{*}$.
For our worked example ($\sigma_{\mathrm{raw}}^{*} = 0.667$, $c = 0.70$):
$M^{*} = 1 + 0.333 \times 0.70 / 0.667 = 1.35$. The corrector makes the
system appear 35\% more competent than it actually is. This gap is the
masking problem in numerical form.

\subsection*{Delegation graphs for uncertainty propagation}

\begin{figure}[htbp]
\centering
\includegraphics[width=\textwidth]{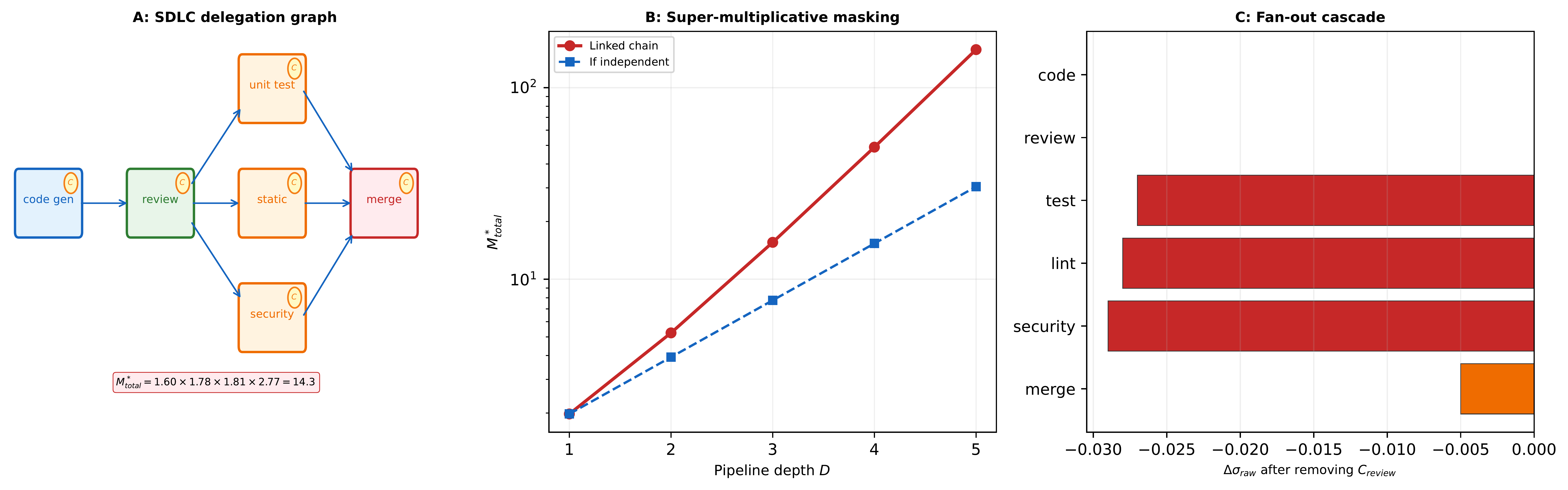}
\caption{The SDLC delegation graph ($\sigma_{\mathrm{skill}} = 0.55$,
$c = 0.65$, product aggregation at merges). \textbf{(A)}~Agent nodes
with correctors; blue arrows show output$\to$input linkage. The masking
index $M^{*}$ increases with depth, reaching $2.77$ at the merge gate
($M^{*}_{\mathrm{total}} = 14.3$). \textbf{(B)}~Total masking compounds
super-multiplicatively with depth: linked chains (solid) grow faster than
the independent prediction (dashed). \textbf{(C)}~Removing the reviewer's
corrector cascades $\Delta\sigma_{\mathrm{raw}}$ to three downstream
branches simultaneously. All values are specific to the stated
parameters.}
\label{fig:delegation}
\end{figure}

Delegated AI workflows are often directed acyclic graphs (DAGs): the code
generator's output feeds an evaluator, whose corrected output fans out to unit
tests, static analysis, and security scanning, all of which fan in to a merge
gate (Figure~\ref{fig:delegation}A). An agent's effective skill depends on the
quality of its inputs:
\begin{equation}\label{eq:effective_skill}
    \sigma_{\mathrm{skill,eff}}(v) = \sigma_{\mathrm{skill}}(v)\;
    \times\; \mathrm{AGG}\bigl(\sigma_{\mathrm{corr}}(u_1),\ldots,
    \sigma_{\mathrm{corr}}(u_k)\bigr),
\end{equation}
where $u_1,\ldots,u_k$ are the parents of $v$ in the DAG and
$\mathrm{AGG}$ is an aggregation function. The choice of $\mathrm{AGG}$
is not arbitrary---it is determined by the task structure at the merge
node:

\begin{itemize}\setlength\itemsep{3pt}
    \item $\mathrm{AGG} = \prod$ (\textbf{product}): each input
    contributes an independent dimension. A merge gate in a code pipeline
    requires both correct logic \emph{and} correct style---errors in
    either degrade the merge. This is the most common case and produces
    the highest masking, because errors from all parents compound
    multiplicatively.

    \item $\mathrm{AGG} = \min$ (\textbf{weakest link}): all inputs
    must be correct for the merge to succeed. A safety-critical system
    that gates on ``all checks pass'' is limited by its weakest
    component. This produces the most fragile fan-in but the most
    ``honest''---masking at the merge is dominated by a single parent.

    \item $\mathrm{AGG} = \text{weighted mean}$: inputs have different
    importance. A recommendation system that blends product relevance
    (weight 0.6) and user sentiment (weight 0.4) averages quality
    across inputs. This dilutes both errors and masking.
\end{itemize}

\noindent In \textbf{fan-out}, one node's failure
cascades to all children simultaneously. In \textbf{fan-in}, errors from
parents sharing an upstream source are correlated (\emph{diamond
pattern}), creating conditional fragility at the merge node
(quality drops $29\%$ when the shared source fails).
The masking index compounds super-multiplicatively with depth
(Figure~\ref{fig:delegation}B): a five-layer pipeline where the
first layer has $M^{*}=1.35$ and the index \emph{increases} with depth
(to ${\sim}1.5$, $1.7$, $2.0$, $2.3$ at subsequent layers) has
$M^{*}_{\mathrm{total}} = \prod M^{*}_i = 14.3$---exceeding
$(1.35)^5 = 4.5$, the prediction from naively assuming uniform masking.

\subsection*{The Euler--Lagrange Solution}

The MSO~(\ref{eq:mso}) is a constrained optimization over the
governed-delegation field $\alpha(x,t)$. We solve it using the calculus of
variations. Introducing a Lagrange multiplier $\lambda$ for the delivery
constraint, the Lagrangian is:

\begin{tcolorbox}[colback=gray!3, colframe=gray!40!black,
    fonttitle=\bfseries\small, boxrule=0.5pt, arc=2pt,
    left=6pt, right=6pt, top=6pt, bottom=4pt,
    title=The Lagrangian]
\[
    \mathcal{L} = \int_0^T \!\!\int_S \alpha^{2}\,\sqrt{\det g}\;
    \mathrm{d}x\,\mathrm{d}t
    \;-\; \lambda\Bigl(\int_S
    \alpha\,\sigma_{\mathrm{raw}}\,\mathrm{d}x
    - p_{\min}\,|S|\Bigr).
\]
\end{tcolorbox}

\noindent Taking the functional derivative and setting it to zero:
\[
    \frac{\delta\mathcal{L}}{\delta\alpha}
    = 2\,\alpha(x)\,\sqrt{g(x)} - \lambda\,\sigma_{\mathrm{raw}}(x) = 0
    \quad\Longrightarrow\quad
    \alpha^{*}(x) = \frac{\lambda\,\sigma_{\mathrm{raw}}(x)}{2\,\sqrt{g(x)}}.
\]
Since $g(\sigma) = 1/\sigma(1-\sigma)$ for Bernoulli outcomes
(equation~\ref{eq:fisher}), $\sqrt{g} = 1/\sqrt{\sigma(1-\sigma)}$ and:

\begin{tcolorbox}[colback=blue!3, colframe=blue!40!black,
    fonttitle=\bfseries\small, boxrule=0.5pt, arc=2pt,
    left=6pt, right=6pt, top=6pt, bottom=4pt,
    title=The Euler--Lagrange Solution (Water-Filling)]
\begin{equation}\label{eq:el_solution}
    \alpha^{*}(x,t) = \min\!\Bigl(\alpha_{\max}(x,t),\;\;
    \frac{\lambda}{2}\,
    \sigma_{\mathrm{raw}}(x)\,\sqrt{\sigma_{\mathrm{raw}}(x)\,
    (1 - \sigma_{\mathrm{raw}}(x))}\Bigr).
\end{equation}

\emph{In words:} governed delegation is allocated proportionally to the product
$\sigma_{\mathrm{raw}} \cdot \sqrt{\sigma(1-\sigma)}$, which peaks at
intermediate competence ($\sigma \approx 0.75$). Points where the agent
is moderately competent receive the most delegated workload under governance
(highest marginal return of review-supported operation). Very weak points
($\sigma \ll 0.5$) receive less
(review has diminishing returns when most outputs need rework). Very
strong points ($\sigma \to 1$) also receive less (review rarely finds
errors). See Box~1 for a worked example.
\end{tcolorbox}

\noindent The multiplier $\lambda$ is determined by the delivery
constraint: $\int_S \alpha^{*}\,\sigma_{\mathrm{raw}}\,\mathrm{d}x =
p_{\min}\,|S|$. The solution distributes governed workload according to a
water-filling rule on the Fisher manifold, where the ``water level''
$\lambda/2$ is set by the delivery target. The allocation peaks at
intermediate competence ($\sigma \approx 0.75$) and tapers at both
extremes, paralleling Shannon's water-filling for power allocation
across parallel channels \citep{cover2006}.

Three closed-form quantities follow directly from the solution:

\begin{itemize}\setlength\itemsep{4pt}
    \item \textbf{Masking index:} $M^{*} =
    \sigma_{\mathrm{corr}}^{*}/\sigma_{\mathrm{raw}}^{*}$. When $M^{*}
    > 1$, the corrector is hiding agent errors from the authorization
    mechanism. In our worked example, $M^{*} = 1.35$. In the
    experiments of Section~3, $M^{*} \approx 1.8$ in the standard
    single-layer delegation---the corrector makes the agent appear nearly
    twice as competent as it is.
    
    \item \textbf{Corrector capacity threshold:}
    $K/N \geq \max\!\left(0,\,(p_{\min} - \sigma_{\mathrm{raw}}^{*})/[(1 -
    \sigma_{\mathrm{raw}}^{*})\,c]\right)$, where $K$ is the number of outputs
    the corrector reviews per cycle, $N$ is the scope size, and $c$ is
    the corrector's catch rate. For $p_{\min} = 0.80$,
    $\sigma_{\mathrm{raw}}^{*} = 0.55$, $c = 0.65$: the threshold is
    $K/N > 0.25/0.2925 = 0.85$. The corrector must review at least
    85\% of outputs. Below this ratio, the delegation cannot maintain
    quality, regardless of how the governed workload is allocated.
    
    \item \textbf{Convergence time:} $T_{\mathrm{cal}} \approx
    1/(\eta + \delta)$ per cell. Averages over larger scopes have lower
    sampling variance, but the dynamical time constant remains
    $1/(\eta+\delta)$ under the independent-cell model.
\end{itemize}

\noindent Together, these quantities answer the question posed in
Section~1. The masking index diagnoses when the oversight process is
hiding agent weakness. The capacity threshold determines when delegation
can function at all. The convergence time predicts how long calibration
will take. And the Euler--Lagrange solution prescribes exactly how much
authority to grant at each point---no more, no less.

\medskip

\noindent\textbf{When the MSO has no solution.}\; The variational
problem~(\ref{eq:mso}) is infeasible when the required quality target exceeds
the operational ceiling available under the current raw support,
corrector catch rate, and review capacity. In this case, the MSO does not
prescribe ``more authority''---it prescribes \emph{no delegation} under the
current design. The task must be performed by a more capable agent, assigned
more effective correction, or decomposed into subtasks that fall within the
current system's competence. The $\alpha^{*}$ solution
is also degenerate when $\sigma_{\mathrm{raw}}$ is uniform across the
scope (no information heterogeneity): the Fisher volume element is
constant, $\alpha^{*}(x)$ is the same at every point, and the theory
adds no value beyond a uniform policy.

\subsection*{Asynchronous delegation dynamics}

The synchronous Return Operator is a mean-field approximation. Real
pipelines are asynchronous: agents process at different rates, correctors
queue, events arrive stochastically, and batch refreshes occur on schedules.
A Generalized Stochastic Petri Net (GSPN) \citep{marsan1984} gives the
corresponding concurrency model. In that setting the local support dynamics
become
\begin{equation}\label{eq:gspn_sigma}
    \frac{\partial\sigma_{\mathrm{raw}}(v)}{\partial t}
    = \lambda_{\mathrm{obs}}(v,\mathbf{m})
    [\sigma_{\mathrm{skill,eff}}(v)-\sigma_{\mathrm{raw}}(v)]
    - \lambda_{\mathrm{forget}}(v)
    [\sigma_{\mathrm{raw}}(v)-\sigma_0(v)],
\end{equation}
where $\mathbf{m}$ is the marking state of the net. When the corrector pool is
uncongested, $\lambda_{\mathrm{obs}}\approx \eta$ and
equation~(\ref{eq:gspn_sigma}) reduces to the synchronous model. Under
congestion, calibration slows and throughput, utilization, and queue lengths
become governance observables. The GSPN is therefore the operational model;
the ODE above is its tractable mean-field projection.

\subsection*{Topology as a governance object}

The delegation DAG is not just a wiring diagram; it changes where governance effort has leverage.
Different motifs generate different failure surfaces and therefore different design prescriptions.
Table~\ref{tab:motifs} summarizes the four recurring cases used throughout the paper.

\begin{table}[htbp]
\centering\small
\caption{Delegation motifs as governance objects.}
\label{tab:motifs}
\begin{tabular}{@{}p{0.14\textwidth}p{0.2\textwidth}p{0.26\textwidth}p{0.28\textwidth}@{}}
\toprule
\textbf{Motif} & \textbf{Failure surface} & \textbf{What grows} & \textbf{Governance implication} \\
\midrule
Chain & Layer-by-layer attenuation & Masking and quality loss accumulate with depth & Improve upstream quality first; each stabilized layer protects all downstream layers. \\
Fan-out & Shared upstream error propagation & One node's failure contaminates multiple branches simultaneously & Prioritize high out-degree nodes because one correction protects multiple children. \\
Diamond & Correlated parent errors through a shared source & Average quality may look stable while conditional fragility spikes & Correct the shared upstream source rather than only the merge gate; diagnose by conditional failure, not averages. \\
Merge / min-cut & Weakest-link or synchronized dependency & Overall throughput and autonomy are limited by the bottleneck path & Invest where $\partial C_{\mathrm{op}}/\partial c(v)$ or $\partial T^{*}_{\mathrm{auto}}/\partial c(v)$ is largest, not where local quality is merely lowest. \\
\bottomrule
\end{tabular}
\end{table}

In this sense, topology enters the theory twice: it determines how errors propagate, and it determines where marginal governance investment produces the largest increase in sustainable autonomy.

\subsection*{Delegation capacity}

The MSO determines the optimal governed-delegation allocation for a given pipeline and
task distribution. But what is the \emph{best possible} quality the
pipeline can achieve, optimized over all task distributions? This
ceiling is the \textbf{delegation channel capacity}.

We define it in two related ways, connected through a monotone bridge
under the binary symmetric model. The \emph{primary} definition is
operational: it is the supremum of achievable shipped output quality:

\begin{tcolorbox}[colback=blue!3, colframe=blue!40!black,
    fonttitle=\bfseries\small, boxrule=0.5pt, arc=2pt,
    left=6pt, right=6pt, top=6pt, bottom=4pt,
    title=Delegation Channel Capacity]
\begin{equation}\label{eq:channel_capacity}
    C_{\mathrm{op}}(G, K, B)
    = \sup_{p(\mathrm{task})}\; q^{*}(\mathrm{output}).
\end{equation}

\emph{In words:} the best possible output quality the pipeline can
achieve. No operational policy can exceed $C_{\mathrm{op}}$.
\end{tcolorbox}

\noindent Here $q^{*}(\mathrm{output})$ denotes the quality of the output
that is actually shipped at the sink. In an uncorrected output stage,
$q^{*}=\sigma^{*}_{\mathrm{raw}}$; in a reviewed output stage,
$q^{*}=\sigma^{*}_{\mathrm{corr}}$. The raw signal remains the preferred
authorization signal because it is less affected by masking.

\noindent The \emph{information-theoretic} definition models each node
as a Binary Symmetric Channel (BSC) with effective error rate
$\varepsilon_{\mathrm{eff}}(v) = (1 - \sigma_{\mathrm{skill}}(v))(1 -
c(v))$, giving node capacity $C_{v}^{\mathrm{BSC}} = 1 -
H_{b}(\varepsilon_{\mathrm{eff}}(v))$ bits. The two definitions are
related by a monotone bridge: since $q^{*} = 1 -
\varepsilon_{\mathrm{eff}}$ and $H_{b}$ is monotone on $[0, 0.5]$,
maximizing $q^{*}$ is equivalent to maximizing
$C^{\mathrm{BSC}}$. We use the operational definition
($C_{\mathrm{op}}$ as a quality ceiling) throughout the design and numerical sections,
and invoke the BSC formulation only in the proofs of Theorem~1 and Proposition~2
where information-theoretic tools are needed.

For a single uncorrected node under the conservative $\sigma_0=0$ convention,
the raw-support ceiling is $\eta/(\eta + \delta)$, achieved when
$\sigma_{\mathrm{skill}} = 1$; with output correction, the shipped-quality
ceiling is the corresponding $\sigma^{*}_{\mathrm{corr}}$. For a linear chain
of depth $D$, each layer receives the corrected output of the previous layer.
Define the
\emph{recursive chain quality} by:
\begin{equation}\label{eq:capacity_chain}
    \sigma^{*}_{\mathrm{corr}}(i) = R\bigl(\sigma_{\mathrm{skill}}
    \times \sigma^{*}_{\mathrm{corr}}(i{-}1)\bigr), \qquad
    \sigma^{*}_{\mathrm{corr}}(0) = 1,
\end{equation}
where $R(\cdot)$ is the Return Operator (equation~\ref{eq:fixed_point}).
The operational quality ceiling at depth $D$ is
$C_{\mathrm{op}}(D) = \sigma^{*}_{\mathrm{corr}}(D)$.

This recursive formula accounts for the fact that each corrector
stabilizes the signal before passing it downstream. The simpler
product formula $C_{\mathrm{prod}}(D) = [\sigma^{*}_{\mathrm{corr}}
(1)]^{D}$ treats each layer as if it had the same input quality,
producing a conservative lower bound (see Experiment~6). We use the
recursive formula in the numerical experiments (gap $< 0.002$ vs.\
simulation) and the information-theoretic bound
$C_{\mathrm{chain}}^{\mathrm{BSC}} = \min_{i}[1 -
H_{b}(\varepsilon_{\mathrm{eff}}(i))]$ in the proofs (where the DPI
is needed).

The capacity has a striking practical consequence. If the quality
requirement $p_{\min}$ exceeds $C_{\mathrm{op}}$, \emph{no
governance policy within the existing pipeline can help}. The designer
must change the pipeline: better agents, more correctors, or a
different topology.

\medskip

The dual quantity is the \textbf{governance overhead}: the minimum cost
of achieving a given quality target.
\[
    R(p_{\min}) = \min_{\pi:\, \sigma^{*}_{\mathrm{raw}} \geq p_{\min}}
    \mathrm{Cost}(\pi),
\]
where $\pi$ ranges over all governance policies and
$\mathrm{Cost}(\pi)$ is the total corrector budget plus routing
overhead. The duality $R(p_{\min}) \leq \mathrm{Budget}
\Leftrightarrow p_{\min} \leq C_{\mathrm{op}}(\mathrm{Budget})$
connects the two quantities.

\medskip

\noindent\textbf{The governed delegation channel.}\;

To prove a capacity theorem, we need a precise channel model where
governance is not an analogy to coding but a \emph{causal action
policy} over a controlled discrete memoryless channel (DMC).

\begin{tcolorbox}[colback=gray!3, colframe=gray!40!black,
    fonttitle=\bfseries\small, boxrule=0.5pt, arc=2pt,
    left=6pt, right=6pt, top=6pt, bottom=4pt,
    title=Definition 1: Governed Delegation Channel]
A \textbf{governed delegation channel} at node $v$ is a
cost-constrained action-dependent DMC
$(\mathcal{X}, \mathcal{A}, \mathcal{Y}, P_{Y|X,A}, c, B)$:

\medskip

\emph{Task alphabet:} $\mathcal{X} = \{0, 1\}$ (correct/incorrect
task specification).

\emph{Action alphabet:} $\mathcal{A} = \{0, 1\}$, where $A_t = 0$
(no review) and $A_t = 1$ (review). Each action incurs cost
$c(A_t)$, with $c(0) = 0$ and $c(1) = 1$, subject to an average
budget: $\frac{1}{n}\sum_{t=1}^{n} \mathbb{E}[c(A_t)] \leq B$.

\emph{Output alphabet:} $\mathcal{Y} = \{0, 1\}$ (correct/incorrect
output).

\emph{Channel law:} Conditional on action $A_t = a$, the node
operates as a BSC with crossover probability $\varepsilon_a$:
$P_{Y|X,A=0} = \mathrm{BSC}(\varepsilon_0)$ (unreviewed),
$P_{Y|X,A=1} = \mathrm{BSC}(\varepsilon_1)$ (reviewed), with
$\varepsilon_1 < \varepsilon_0$. For the delegation setting:
$\varepsilon_0 = 1 - \sigma_{\mathrm{skill}}(v)$ and
$\varepsilon_1 = (1 - \sigma_{\mathrm{skill}}(v))(1 - c_v)$.

\emph{Memoryless property:} $P(Y^n | X^n, A^n) = \prod_{t=1}^{n}
P(Y_t | X_t, A_t)$.
\end{tcolorbox}

\noindent\emph{Key distinction:} Governance is not a code applied
\emph{after} the channel---it is a causal action that selects
\emph{which channel law} is applied at each symbol, under a cost
constraint. The corrector's review is the action; the review budget
is the cost constraint.

\medskip

\noindent\textbf{Delegation capacity.}\; In full generality, the capacity of
a governed delegation channel is the supremum of achievable rates under
admissible causal governance policies:
\begin{equation}\label{eq:del_capacity}
    C_{\mathrm{del}}(B)
    = \sup_{\pi \in \Pi(B)}\;
    \liminf_{n \to \infty} \frac{1}{n}\,I(X^n; Y^n),
\end{equation}
where $\Pi(B)$ is the set of causal governance policies satisfying
the average review-cost budget $B$. The fully adaptive feedback case is a
controlled-channel problem and is not solved in closed form here. The theorem
below proves the single-letter capacity for the stationary symbolwise case
used in the simulations and design calculations.

\noindent\textbf{Single-letter form (stationary symbolwise policies).}\;
When the governance policy is restricted to stationary symbolwise
review---i.e., each task is independently reviewed with probability
$q$, without dependence on past observations---and the review action is
recorded in the governance log, the effective channel output is the pair
$(Y,A)$. The action is a Bernoulli random variable
$A \sim \mathrm{Bern}(q)$ with $q \leq B$ (review fraction), independent
of $X$. Each action induces a BSC, so
$I(X;Y,A)=I(X;Y\mid A)$ and
$I(X;Y \mid A = a) = 1 - H_b(\varepsilon_a)$ for
$X \sim \mathrm{Bern}(1/2)$. Optimizing over admissible $q$ gives the
revealed-action capacity:
\begin{equation}\label{eq:single_letter}
    C_{\mathrm{del}}(B) = \max_{q \in [0,B]}\;
    (1-q)\bigl[1 - H_b(\varepsilon_0)\bigr]
    + q\bigl[1 - H_b(\varepsilon_1)\bigr].
\end{equation}
Since $H_b(\varepsilon_1) < H_b(\varepsilon_0)$ for
$\varepsilon_1 < \varepsilon_0 < 1/2$, the optimum is $q^{*} = B$
(review as much as the budget allows), giving:
\[
    C_{\mathrm{del}}(B) = (1-B)\bigl[1 - H_b(\varepsilon_0)\bigr]
    + B\bigl[1 - H_b(\varepsilon_1)\bigr].
\]
If the action log is not available to the decoder or evaluator, the
stationary randomized policy induces the averaged BSC with crossover
$\varepsilon_{\mathrm{avg}}(q)=(1-q)\varepsilon_0+q\varepsilon_1$ and
capacity $1-H_b(\varepsilon_{\mathrm{avg}}(q))$, which is generally
different. The paper uses the revealed-action form because governance
actions are logged by construction.

\begin{tcolorbox}[colback=green!3, colframe=green!40!black,
    fonttitle=\bfseries\small, boxrule=0.5pt, arc=2pt,
    left=6pt, right=6pt, top=6pt, bottom=4pt,
    title={Theorem 1 \normalfont{[Theorem]}: Delegation Capacity}]
\emph{Assumptions:} Each node $v$ operates as a governed delegation
channel (Definition~1). Review actions are stationary, symbolwise,
independent of $X$, and recorded; the decoder/evaluator observes the
action log. Nodes process independently conditional on their input.
Tasks are processed in epochs of $n$ symbols, with
$0 \leq \varepsilon_1 < \varepsilon_0 < 1/2$.

\medskip

\emph{Achievability:} For any source with entropy rate
$H(X) < C_{\mathrm{del}}(B)$, there exists an encoder, a governance
policy satisfying the stationary symbolwise budget, and a decoder such that the block error
probability $P_e^{(n)} \to 0$ as $n \to \infty$.

\medskip

\emph{Converse:} If $H(X) > C_{\mathrm{del}}(B)$, then for every
encoder, stationary symbolwise admissible governance policy, and decoder:
$\liminf_{n \to \infty} P_e^{(n)} > 0$.

\medskip

For a cascade of $D$ governed delegation channels:
$C_{\mathrm{del}}^{\mathrm{chain}}(B_1, \ldots, B_D) \leq \min_{v}
C_{v}(B_v)$, with equality in the binary memoryless cascade under
end-to-end coding across the composed channel.
\end{tcolorbox}

\noindent\emph{Proof of achievability.}\;

\emph{Step~A (fix a stationary governance law).}\; Choose a
stationary action law $P_A = \mathrm{Bern}(q)$ with $q \leq B$. Because
the action is recorded, the channel output is $(Y,A)$ and the channel law
factorizes as:
\[
    P_{Y,A|X}^{(q)}(y,a|x) = P_A(a)\,P_{Y|X,A=a}(y|x).
\]
The mutual information is
$I(X;Y,A)=I(X;Y\mid A)$ because $A$ is independent of $X$.

\emph{Step~B (standard random coding).}\; For any rate
$R < I(X;Y,A)_{P_X, P_A, P_{Y|X,A}}$, standard random coding on
the induced DMC $P_{Y,A|X}^{(q)}$ produces a sequence of
$(2^{nR}, n)$ block codes with $P_e^{(n)} \leq 2^{-n E(R)}$, where
$E(R) > 0$ is the reliability function of the induced DMC
\citep{cover2006}, Theorem~7.7.1.

\emph{Step~C (optimize over governance).}\; Take the supremum over
admissible $q \in [0,B]$. The maximum is attained at $q^{*} = B$,
yielding $C_{\mathrm{del}}(B)$ as in
equation~(\ref{eq:single_letter}).

\medskip

\noindent\emph{Proof of converse.}\; Let $\hat{X}^n$ be the decoded
task sequence after the full pipeline. If $P_e^{(n)} \to 0$, Fano's
inequality \citep{cover2006}, Theorem~2.10.1, gives:
\[
    nR = H(X^n) \leq I(X^n; \hat{X}^n) + n\epsilon_n,
    \quad \epsilon_n \to 0.
\]
For the pipeline Markov chain with recorded governance actions
$X^n \to (Y_1^n,A_1^n) \to (Y_2^n,A_2^n) \to \cdots
\to (Y_D^n,A_D^n) \to \hat{X}^n$,
the DPI gives:
\[
    I(X^n; \hat{X}^n) \leq \min_{v} I(X^n; Y_v^n,A_v^n).
\]
For each node $v$, under admissible action budget $B_v$:
$\frac{1}{n} I(X^n; Y_v^n,A_v^n) \leq C_v(B_v)$. Therefore
$R \leq \min_v C_v(B_v)$.\hfill$\square$

\medskip

\noindent\textbf{Corollary (quality target).}\; The theorem is
stated in bits. To translate to the operational quality target
$p_{\min}$: for a uniform binary source, mutual information is
$I(X;Y) = 1 - H_b(1-p)$ where $p$ is the symbol correctness
probability. Therefore, a quality target $p_{\min}$ is achievable if
and only if:
\[
    1 - H_b(1 - p_{\min}) < C_{\mathrm{del}}(B).
\]
Equivalently: $p_{\min} < 1 - H_b^{-1}(1 - C_{\mathrm{del}}(B))$.
This is the bridge between the information-theoretic capacity (in
bits) and the operational quality ceiling (in $\sigma$) used
throughout the paper.

\medskip

\noindent\emph{Remark (DAGs).}\; For a general DAG, the chain
bottleneck generalizes to a cut-set bound:
\[
    R \;\leq\; \min_{\mathcal{C} \in \mathrm{cuts}(G)} \sum_{v \in \mathcal{C}} C_v(B_v),
\]
under the conditional independence assumptions needed for network
factorization \citep{cover2006}, the network-information-theory
cut-set viewpoint of \citet[Section~15.10]{cover2006}. When upstream errors are correlated (diamond pattern),
the effective capacity is lower in theory; in the mean-field simulator,
the average-quality penalty is negligible ($< 1\%$), but conditional
analysis reveals a $1.4\times$ fragility ratio: quality at $D$ drops
$29\%$ when the shared source $A$ fails (Section~3).

\medskip

\noindent\emph{Remark (Fisher prioritization).}\; The Fisher-priority
review rule ($\phi_{\pi}(v) = g(\sigma_{\mathrm{raw}}(v))$) is not
part of the capacity theorem. It is a separate result: under a
quadratic local loss, Fisher-prioritized review maximizes the
first-order marginal gain in mutual information per unit review cost.
The practical advantage of Fisher priority over uniform review depends
on the heterogeneity of $\sigma$ across scope points; in settings with
homogeneous competence, the improvement is negligible.
This belongs as an operational prescription (Section~4), not as the
constructive policy in the achievability proof.

\subsection*{Process entropy and process capacity}

The delegation capacity bounds quality for a given pipeline. But real
workflows vary in \emph{complexity}---the number of decisions agents
make, the stochasticity of routing, the asynchrony of events. A
deterministic pipeline (every item follows the same path) is easier to
govern than a stochastic one (agents choose tools and routes on the fly).

We define the \textbf{process entropy} of a workflow $W$ as the total
Shannon entropy of the execution decisions:
\begin{equation}\label{eq:process_entropy}
    H(W) = H(\text{routing}) + H(\text{tool calls}) + H(\text{timing}),
\end{equation}
where each term is defined precisely:
$H(\text{routing}) = \sum_{v} H(P_v)$, where $P_v$ is the probability
distribution over outgoing edges at node $v$ (deterministic routing has
$H(P_v) = 0$; an agent that chooses among $k$ routes uniformly has
$H(P_v) = \log_2 k$); $H(\text{tool calls}) = \sum_{v} H(Q_v)$, where
$Q_v$ is the distribution over tool-call sequences at node $v$; and
$H(\text{timing}) = \sum_{v} H(T_v)$, where $T_v$ is the distribution
over inter-event arrival times at node $v$ (synchronous steps have
$H(T_v) = 0$). The additive decomposition assumes that routing, tool,
and timing decisions at different nodes are \emph{conditionally
independent given the input}---a natural assumption when agents do not
share hidden state.

The following result connects capacity to workflow complexity. Unlike
Theorem~1, which is proved for the revealed-action stationary governed
delegation channel,
Proposition~2 relies on a Taylor approximation and an
entropy-variance bound. It should be read as a \emph{local first-order sensitivity law}:
near a reference workflow regime, added process complexity acts as an approximately
linear tax on achievable quality. It captures the empirically observed linear
degradation (Section~3), but is not intended as a globally tight bound:

\begin{tcolorbox}[colback=green!3, colframe=green!40!black,
    fonttitle=\bfseries\small, boxrule=0.5pt, arc=2pt,
    left=6pt, right=6pt, top=6pt, bottom=4pt,
    title={Proposition 2 \normalfont{[Proposition]}: Local Process-Complexity Sensitivity Law}]
For a pipeline operating near a reference workflow regime, with channel capacity $C_{\mathrm{op}}$ and
process entropy $H(W)$ measured relative to that regime:
\begin{equation}\label{eq:process_capacity}
    \sigma^{*}_{\mathrm{raw}}(\mathrm{output} \mid W)
    \;\geq\; C_{\mathrm{op}} - \lambda\, H(W),
\end{equation}
where $\lambda$ is the \textbf{governance gap coefficient}.

\medskip

\emph{In words:} near a fixed operating regime, each additional bit of workflow complexity
reduces achievable output quality by approximately $\lambda$. This is a local sensitivity,
not a universal law: it says how fast quality erodes around the current regime, not how every
workflow behaves globally.
\end{tcolorbox}

\noindent\emph{Proof.}\; Model the routing state as a random variable
$R$ with entropy $H(R) = H(W)$, so the delegation channel's transition
matrix $p(Y|X)$ depends on $R$. Consider a local perturbation around a reference routing
regime $r_0$ where the conditional capacity is twice differentiable. Using the standard chain-rule identity
$ I(X;Y,R) = I(X;R) + I(X;Y \mid R) = I(X;Y) + I(X;R \mid Y)$,
we obtain
\[
    I(X;Y) = I(X;Y \mid R) + I(X;R) - I(X;R \mid Y).
\]
For exogenous workflow randomness, $R$ is independent of the source $X$,
so $I(X;R)=0$ and therefore $I(X;Y) \le I(X;Y \mid R)$. Averaging over
routing realizations gives
\[
    I(X;Y) \leq \mathbb{E}_{R}[I(X;Y \mid R)] \leq C_{\mathrm{op}},
\]
where the second inequality uses Fano's inequality with side information
\citep{cover2006}: for each realization $r$, the achievable quality is
bounded by the channel capacity, $I(X;Y \mid R=r) \le C_{\mathrm{op}}$.
The gap between $I(X;Y)$ and $C_{\mathrm{op}}$ arises because the
receiver (corrector) must allocate review effort across all possible
routing realizations rather than the optimal allocation for the actual
realization. By a second-order Taylor expansion of the conditional capacity around
the reference routing regime, the first-order term vanishes at the local optimum and the residual gap is bounded by:
\[
    C_{\mathrm{op}} - I(X;Y) \leq
    \frac{1}{2}\,\max_{r}\!\bigl|C''(r)\bigr|\;
    \mathrm{Var}(R) \leq \lambda\, H(R),
\]
where the last inequality should be read as a local normalized
entropy--dispersion bound: for the finite routing alphabets used here, the
routing coordinate is scaled to bounded support and the proportionality
constant between dispersion and entropy is absorbed into $\lambda$. Thus
$\lambda$ is a local property of the channel and routing parametrization, not
a universal constant.
Since $\sigma^{*}_{\mathrm{raw}}$ is a monotone function of $I(X;Y)$,
the bound transfers: $\sigma^{*}_{\mathrm{raw}} \geq C_{\mathrm{op}}
- \lambda H(W)$.

\medskip

\noindent\emph{Worked example:} A pipeline with $C = 0.80$,
$p_{\min} = 0.50$, $\lambda = 0.02$/bit can handle workflows with up to
$H_{\max} = (0.80 - 0.50)/0.02 = 15$ bits of process entropy---roughly
15 independent binary routing decisions, or $2^{15}$ equally likely routing
traces---before quality drops below threshold.

\emph{Remark.}\; The governance gap $\lambda$ is now a \emph{derived}
quantity---the curvature of the conditional capacity as a function of
routing state---though in practice it is easier to measure empirically
(Experiment~7 reports $\lambda \approx 0.02$/bit). Better governance
(Fisher-prioritized routing) reduces $\lambda$ by concentrating review
on the most uncertain outcomes, partially compensating for routing
uncertainty.

\medskip

\noindent The maximum process complexity at quality $p_{\min}$ is
therefore:
\[
    H_{\max}(p_{\min})
    = \frac{C_{\mathrm{op}} - p_{\min}}{\lambda}.
\]
This is the \textbf{delegation process capacity}: the maximum entropy
of a workflow that the pipeline can handle while maintaining quality.
Better governance (higher $K/N$, Fisher-prioritized routing) reduces
$\lambda$, allowing more complex workflows at the same quality.

It is useful to summarize these constraints by a single \textbf{effective
autonomy buffer}:
\begin{equation}\label{eq:effective_buffer}
    B_{\mathrm{eff}} = C_{\mathrm{op}} - p_{\min} - \lambda H(W).
\end{equation}
This buffer is the central geometric quantity of the theory. When
$B_{\mathrm{eff}} > 0$, delegated autonomy is feasible under the model.
When $B_{\mathrm{eff}} = 0$, the pipeline is exactly at its autonomy
cliff. When $B_{\mathrm{eff}} < 0$, no governance policy can sustain
the requested quality target. The autonomy time in the next subsection
is the temporal survivability of this same buffer under drift and noise.
Table~\ref{tab:shannon} summarizes the structural correspondence
between Shannon's information theory and the delegation framework.
Note: this is a \emph{structural} correspondence, not a formal
equivalence. The formal theorem (Theorem~1) is stated for the
revealed-action stationary governed delegation channel, not by analogy.

\begin{table}[htbp]
\centering\small
\caption{Structural correspondence between information theory and
delegation theory (see Theorem~1 for the formal result).}
\label{tab:shannon}
\begin{tabular}{@{}ll@{}}
\toprule
\textbf{Information theory} & \textbf{Delegation theory} \\
\midrule
Channel $P_{Y|X}$ & Governed delegation channel $P_{Y|X,A}$ \\
Channel capacity $\max I(X;Y)$ &
    Delegation capacity $C_{\mathrm{del}}(B)$ \\
Error-correcting code &
    Governance policy $\pi = (K, \rho, \phi)$ \\
Source entropy $H(X)$ & Process entropy $H(W)$ \\
\bottomrule
\end{tabular}
\end{table}

\subsection*{Autonomy time}

The most operationally relevant quantity is the \textbf{autonomy time}
$T^{*}_{\mathrm{auto}}$: the expected duration a pipeline can operate
without human intervention before quality drops below $p_{\min}$.

In the GSPN framework, the system's state drifts due to distribution
shift (agents' effective skill degrades at rate $\mu$) and fluctuates
due to finite sampling (noise variance $\nu^{2}$). We model the output
quality as a stochastic process:
\[
    \mathrm{d}\sigma(t) = -\mu_{\mathrm{eff}}\,\mathrm{d}t
    + \nu_{\mathrm{eff}}\,\mathrm{d}W(t),
\]
where $W(t)$ is a standard Wiener process. The autonomy time is the
\emph{first-passage time} of $\sigma(\text{output},t)$ below $p_{\min}$,
starting from the effective operating point $\sigma_{0}$, so that the initial distance to the intervention threshold is precisely the effective autonomy buffer $B_{\mathrm{eff}}$ from equation~\ref{eq:effective_buffer}.

For this drifted Brownian motion, in the ideal constant-drift model with
$\mu_{\mathrm{eff}}>0$, the mean first-passage time from $\sigma_{0}$ to
the absorbing barrier at $p_{\min}$ is (see e.g.\ \citealt{redner2001}):

\begin{tcolorbox}[colback=blue!3, colframe=blue!40!black,
    fonttitle=\bfseries\small, boxrule=0.5pt, arc=2pt,
    left=6pt, right=6pt, top=6pt, bottom=4pt,
    title={Proposition 3 \normalfont{[Proposition]}: Autonomy Time}]
\begin{equation}\label{eq:autonomy_time}
    T^{*}_{\mathrm{auto}}
    = \frac{C_{\mathrm{op}} - p_{\min}
    - \lambda\,H(W)}{\mu_{\mathrm{eff}}}.
\end{equation}

\emph{In words:} the maximum expected time the pipeline can operate
without human intervention. The numerator is the effective autonomy buffer from
equation~\ref{eq:effective_buffer}; the denominator is
the drift rate. Autonomy is proportional to the buffer and inversely
proportional to the drift.
\end{tcolorbox}

\noindent In the ideal Brownian model with constant negative drift, the
mean is $B_{\mathrm{eff}}/\mu_{\mathrm{eff}}$ and the diffusion term
controls the dispersion of hitting times rather than the mean. When drift is
negligible and noise dominates, the mean first-passage time of an unbounded
Brownian motion to a one-sided barrier is not finite; nevertheless, typical
times such as medians and fixed quantiles scale as
$(\sigma_{0}-p_{\min})^{2}/\nu^{2}_{\mathrm{eff}}$. The experiments
(Section~3) are therefore interpreted as testing the drift-dominated scaling
$T^{*}_{\mathrm{auto}} \propto 1/\mu$, not as a full distributional
first-passage theorem.

\noindent\textbf{Five factors of autonomy.}\; $T^{*}_{\mathrm{auto}}$
increases when:
\begin{enumerate}\setlength\itemsep{1pt}
    \item $C_{\mathrm{op}}$ is high (better agents, correctors,
    topology).
    \item $p_{\min}$ is low (relaxed quality requirements).
    \item $H(W)$ is low (simpler, more deterministic workflows).
    \item $\lambda$ is small (better governance compresses the gap).
    \item $\mu_{\mathrm{eff}}$ and $\nu^{2}_{\mathrm{eff}}$ are small
    (stable models, large scope, frequent observations).
\end{enumerate}

\noindent The minimum human intervention frequency at each node is
$f(v) = 1/T^{*}_{\mathrm{auto}}(v)$, and the optimal intervention
schedule minimizes total human review cost:
\[
    \min \sum_{v} f(v) \times \mathrm{cost}(v)
    \quad\text{subject to}\quad
    f(v) \geq 1/T^{*}_{\mathrm{auto}}(v)\;\;\forall\, v.
\]
This is a linear program, solvable in polynomial time. Nodes with
short $T^{*}_{\mathrm{auto}}$ (high drift, high process entropy, low
capacity) receive more frequent human review; nodes with long
$T^{*}_{\mathrm{auto}}$ can operate autonomously for extended periods.

The convergence time for the pipeline to reach an $\epsilon$-neighborhood
of operational capacity after deployment (or after a modification) is:
\[
    T_{\mathrm{cal}} = \frac{\ln(1/\epsilon)}{\lambda_{1}},
\]
where $\lambda_{1}$ is the spectral gap of the coupled GSPN generator
matrix. The expected number of observations collected over that interval is
approximately $N\bar{\eta}T_{\mathrm{cal}}$, where $\bar{\eta}$ is the
average observation rate per scope point. Fan-out
accelerates convergence (parallel observations); resource contention
decelerates it (queuing delays).

\section{Connections to Existing Frameworks}

The MSO connects to established frameworks in economics and security.
Table~\ref{tab:unification} summarizes the correspondences; we develop
the two strongest below.

\begin{table}[htbp]
\centering\small
\caption{Correspondences between established frameworks and MSO
constructs.}
\label{tab:unification}
\begin{tabular}{@{}lll@{}}
\toprule
\textbf{Existing concept} & \textbf{Field} & \textbf{MSO construct} \\
\midrule
Contract / incentive scheme & Economics & Governance functional $G$ \\
Monitoring signal & Economics & Evidential support $\sigma_{\mathrm{raw}}$ \\
Moral hazard & Economics & Masking problem ($M^{*} > 1$) \\
Access-control policy & Security & Authorization field $\alpha^{*}$ \\
Least privilege & Security & MSO (variational formulation) \\
\bottomrule
\end{tabular}
\end{table}

\subsection*{Principal--agent theory}

In the standard principal--agent model \citep{jensen1976,holmstrom1979},
the principal designs a contract $w(y)$ mapping observable output to
payment, under moral hazard: the agent's effort is unobservable. The
MSO produces a correspondence: the contract $w(y)$ maps to the
governance functional $G(\sigma)$; the output $y$ maps to
$\sigma_{\mathrm{raw}}$; moral hazard maps to masking ($M^{*} > 1$).
The mechanism collapses to a single monotone function $G$ because the
agent's ``type'' is revealed by outcomes the corrector observes---no
message space or revelation principle is needed \citep{myerson1981}.

The MSO also identifies a form of moral hazard absent from classical
theory: the information asymmetry created by the \emph{corrector}, not
the agent. The dual~$\sigma$ prescription resolves this by changing how
information is \emph{recorded}, not how agents are \emph{rewarded}.
In a complementary direction, \citet{fudenberg2025} characterize
optimal information disclosure to a single delegate whose alignment is
uncertain; the MSO addresses the orthogonal problem of allocating
oversight across a \emph{pipeline} of delegates.

\subsection*{The principle of least privilege}

Saltzer and Schroeder's \citeyearpar{saltzer1975} principle---``operate
using the least set of privileges necessary''---has remained a
heuristic for fifty years. The MSO provides its variational
formulation: equation~(\ref{eq:mso}) minimizes total oversight cost subject
to delivery, and the water-filling solution~(\ref{eq:el_solution})
gives the computable policy: $\alpha^{*}(x) = \min(\alpha_{\max}(x),
\lambda\sigma(x)/(2\sqrt{g(x)}))$, determined from the system's own
logs and the quality requirement $p_{\min}$.

\medskip

\noindent\emph{Scope of the correspondences.}\; The strongest links are to
principal--agent theory and least privilege. Other analogies---for example to
BDI architectures or span-of-control theory---are possible, but they are
interpretive rather than theorem-bearing and are not needed for the main
results of the paper.

\subsection*{Boundary conditions of the theory}

The correspondences above are strongest where the paper's own assumptions
are strongest. The current formulation is built on binary outcomes,
memoryless node behavior, conditional independence across nodes, and
product or min aggregation at merges. Several extensions are plausible,
but they are not yet proved here. Continuous or multiclass outcomes would
replace the Bernoulli manifold by a higher-dimensional statistical
manifold and alter both the Fisher volume term and the channel model.
Shared hidden state or correlated latent failures would weaken the
conditional-independence assumptions behind the current DAG analysis.
Adaptive governance policies with feedback beyond stationary symbolwise
review would require extending Theorem~1 from the present governed
channel to a richer controlled channel with memory. The framework is
therefore best read as a principled base case whose objects and limits
are explicit, not as a universal theory of all delegation.


\section*{Related Work}
\addcontentsline{toc}{section}{Related Work}

The MSO draws on and extends work across several communities. We
position it relative to the most relevant strands. The discussion is selective: we emphasize the strongest intellectual neighbors and keep the focus on frameworks that change how delegated autonomy should be designed or measured.

\subsection*{AI safety and scalable oversight}

The masking problem is closely related to the \emph{scalable oversight}
challenge in AI alignment: as AI systems become more capable, human
overseers struggle to evaluate outputs they cannot themselves produce
\citep{amodei2016}. Reinforcement learning from human feedback (RLHF)
uses human preferences as a training signal, but the preference model
can itself drift or be gamed---a form of corrector drift in our
framework. Constitutional AI \citep{bai2022} attempts to reduce
dependence on human feedback by having the model critique its own
outputs, but self-critique is a dual-role delegation: each unit of capacity spent on review is a unit not spent on generation, and the incentive structure biases toward the self-rewarding generative role. The MSO provides a formal criterion---$M^{*}$---for detecting
when the oversight process is masking rather than revealing the agent's
true competence. Recent work on scaling laws for scalable oversight
\citep{engels2025} derives empirical relationships between oversight
success and capability gaps; the MSO provides the principled allocation
theory that such empirical observations currently lack. ALARA-style agent
harness engineering similarly emphasizes practical constraints in portable,
composable multi-agent teams \citep{alara2025}; MSO supplies a quantitative
governance objective for such harnesses.

\subsection*{LLM-as-judge and evaluation}

The practice of using one LLM to evaluate another's output
\citep{zheng2023} is a delegation with an autonomous corrector. Recent
work has documented systematic biases in LLM judges: position bias,
verbosity bias, and self-enhancement bias. In the MSO framework, these
biases manifest as a non-uniform catch rate $c(x)$ across the scope,
producing regions where $M^{*}$ is high (the judge is lenient) and
regions where $M^{*}$ is low (the judge is strict). The dual~$\sigma$
diagnostic detects this heterogeneity: when
$\sigma_{\mathrm{raw}}(x)$ varies across regions while
$\sigma_{\mathrm{corr}}(x)$ does not, the judge is applying different
standards in different parts of the scope.

\subsection*{Delegated AI and agent orchestration}

Research on delegated AI and agent orchestration has long addressed how to
distribute tasks and coordination, from contract nets \citep{smith1980} and
BDI architectures \citep{rao1991} to holonic architectures and classical
span-of-control questions \citep{urwick1956}. What those frameworks usually lack is a governing
optimality principle for autonomy allocation itself. The MSO supplies such a
principle and turns heuristic questions---where to place evaluation, how to
size oversight, when to decompose scope---into computable quantities.

\subsection*{Agentic AI frameworks and orchestration}

Recent orchestration frameworks make delegated pipelines easy to build:
AutoGen \citep{autogen2023}, CrewAI \citep{crewai2024}, LangGraph
\citep{langgraph2024}, Google ADK \citep{adk2025}, and task-adaptive
orchestration systems such as AdaptOrch \citep{adaptorch2025}. They do not by
themselves specify how much autonomy each node should receive, where
correctors belong, or when intervention becomes necessary.
\citet{tomasev2026} propose a comprehensive conceptual framework for
AI delegation covering sub-delegation, permission attenuation, and
reputation systems, but explicitly without formal bounds or theorems. Static
workflow verification \citep{agentproof2025} and resource-bounded agent
contracts \citep{agentcontracts2025} are complementary: they constrain or
verify agent graphs, while the MSO allocates autonomy and oversight over those
graphs.
The MSO is best read as a layer above such frameworks: it treats their
graphs as governance objects and supplies the formal quantities---capacity
ceilings, masking diagnostics, intervention timing---that conceptual
and orchestration frameworks currently lack.

\subsection*{Variational and information-processing viewpoints}

Variational principles also appear in predictive-processing and free-energy
formulations \citep{friston2010}. The MSO uses a variational form in a narrower
engineering sense: minimize a governance functional subject to a delivery
constraint. It does not adopt the free-energy principle as a cognitive or
biological theory.

\subsection*{Information geometry}

The Fisher information metric is the paper's main geometric commitment.
Where information geometry usually appears in learning or estimation,
including natural-gradient learning \citep{amari1998}, we use it to allocate
governance attention across a delegated task space.
We do not claim a full information-geometric theory of delegated control;
we use the Fisher geometry because it is the natural metric for the local
cost of calibration.

\subsection*{Computational considerations}

The Euler--Lagrange solution~(\ref{eq:el_solution}) is computable in
$O(N)$ time per scope update, where $N$ is the number of scope points:
it requires computing $\sigma_{\mathrm{raw}}$ at each point (a running
average, $O(1)$ per observation) and the Fisher volume integral (a sum
over the scope, $O(N)$). The masking index $M^{*}$ is a single division
per node. For delegation DAGs, the principled improvement target is the
local sensitivity $\partial T^{*}_{\mathrm{auto}}/\partial c(v)$ or its
finite-difference approximation; when exact sensitivities are not
available, the proxy score $S(v) = \mathrm{DC}(v) \times M^{*}(v)
\times \kappa(v)$ remains computable after a single topological sort
($O(|V| + |E|)$) to evaluate $\mathrm{DC}$ for all nodes. The diagnostic
differential $(\Delta\sigma_{\mathrm{raw}}, \Delta M^{*})$ is a
comparison of two consecutive windows. The theory is therefore tractable enough to support production monitoring and iterative redesign, not just offline analysis.

\subsection*{Identifiability and measurement}

The theory is only useful operationally if its main quantities can be
estimated from logs. The distinction among what is directly observed,
what is inferred, and what is only approximated is therefore part of the
framework rather than an implementation detail. Corrected output quality
$\sigma_{\mathrm{corr}}$ is usually directly observable from post-review
or post-execution outcomes. Raw competence $\sigma_{\mathrm{raw}}$ is
harder: it requires either a pre-correction slice, a shadow evaluation
channel, or a model-based estimator that reconstructs the counterfactual
uncorrected outcome stream. Catch rate $c$ is not usually observed as a
primitive quantity; it is inferred from reviewed-error statistics and is
therefore vulnerable to evaluator drift. Process entropy $H(W)$ is
estimated from routing, tool, and timing traces, and its quality depends
on the completeness of those traces. Drift and noise terms
$\mu_{\mathrm{eff}}$ and $\nu_{\mathrm{eff}}$ are fitted from temporal
windows and inherit the usual local-stationarity assumptions.

\begin{table}[htbp]
\centering
\small
\begin{tabular}{p{2.2cm}p{1.7cm}p{3.0cm}p{3.1cm}p{3.0cm}}
\toprule
Quantity & Observable? & Typical estimator & Main assumption & Main failure mode \\
\midrule
$\sigma_{\mathrm{corr}}$ & Yes & Post-review success rate & Outcome logging is correct & Hidden rerouting or silent overrides \\
$\sigma_{\mathrm{raw}}$ & Partial & Pre-review slice / shadow channel / synthetic estimator & Counterfactual pre-correction stream is identifiable & Leakage between corrected and raw channels \\
$c$ & No (inferred) & Reviewed-error catch fraction & Reviewed errors are representative & Corrector drift or selective review bias \\
$H(W)$ & Approx. & Entropy of routing/tool/timing traces & Trace coverage is complete enough & Hidden state or missing tool events \\
$\mu_{\mathrm{eff}}, \nu_{\mathrm{eff}}$ & No (fitted) & Drift/noise fit over rolling windows & Local stationarity & Regime shift or short windows \\
\bottomrule
\end{tabular}
\caption{Measurement status of the main quantities in the framework. The point is not that every quantity is directly observed, but that the observational status of each is explicit.}
\label{tab:measurement}
\end{table}

This perspective also clarifies why dual tracking matters. The theory
requires two informational channels because raw competence and delivered
quality are different observables. Conflating them not only creates the
masking pathology; it also destroys identifiability of the variable that
the governance law actually needs.


\section{Numerical Validation}

All numerical results are reproducible via the companion package
\texttt{minimal-oversight} \citep{delegationlab2026}. The simulator uses a
mean-field ODE model with Bernoulli outcome noise over $N=80$--$144$ scope
points. Unless otherwise stated, $\eta=10$, $\delta=2$,
$\sigma_{\mathrm{skill}}=0.55$--$0.80$, $c=0.65$--$0.70$, and
$K/N=0.50$. Each experiment runs for 200--400 steps over $n=20$--$30$
independent seeds. The corrector model is the expected-value update
\[
    \sigma_{\mathrm{corr}}
    = \sigma_{\mathrm{raw}} + (1-\sigma_{\mathrm{raw}})\,c\,(K/N).
\]
This is not a holdout measurement protocol: in production,
$\sigma_{\mathrm{raw}}$ and $\sigma_{\mathrm{corr}}$ should be measured on
separate pre- and post-correction channels.

\subsection*{Semi-real reconstructed workflow}

We also evaluate a reconstructed software-delivery workflow with 2{,}400
items over four windows. Items pass through a generator, reviewer, three
parallel checks (test, requirements, security), and a merge decision. The
reviewer has the highest masking and upstream leverage:
$\sigma_{\mathrm{raw}}=0.62$, $\sigma_{\mathrm{corr}}=0.90$, and
$M^{*}=1.45$. The workflow entropy is $H(W)=2.3$ bits; the estimated
pipeline ceiling is $C_{\mathrm{op}}\approx 0.86$. For $p_{\min}=0.75$
and $\lambda H(W)\approx0.046$, the effective buffer is
$B_{\mathrm{eff}}\approx0.064$. With $\mu_{\mathrm{eff}}\approx0.012$,
the implied autonomy time is about $5.3$ time units. The design
prescription is upstream: reduce reviewer masking or simplify branch routing
before spending more effort at the merge gate.

\subsection*{Controlled experiments}

Table~\ref{tab:verification} summarizes the controlled validation. The
experiments should be read as internal model checks rather than external
field validation: they test whether the closed-form quantities and scaling
laws match the simulator generated from the same assumptions.

\begin{table}[htbp]
\centering\small
\caption{Summary of controlled validation results from the companion package.}
\label{tab:verification}
\resizebox{\textwidth}{!}{%
\begin{tabular}{@{}p{3.4cm}p{5.3cm}p{5.2cm}@{}}
\toprule
\textbf{Experiment} & \textbf{Prediction tested} & \textbf{Observed result} \\
\midrule
Masking & Corrected quality hides raw competence; single-$\sigma$ policies
over-authorize. & $M^{*}\approx1.4$ during correction; single-$\sigma$
governance over-authorizes by $36\%$. \\
Communication & Routing to the right corrector and prioritizing uncertain
items improves quality. & Strategic routing/Fisher priority improves
$\sigma^{*}$ from $0.545$ to $0.579$ ($+6.1\%$). \\
Bottlenecks & Limited corrector capacity should dominate other isolated
failure modes. & Capacity loss $-0.106$, larger than adversarial drift
($-0.035$), correlations ($-0.008$), or heterogeneity ($-0.002$). \\
Linked chains & Masking increases with depth when corrected outputs feed
downstream layers. & Per-layer $M^{*}$ rises from $1.77$ to $2.20$ over
five layers; total masking is $38.7$. \\
DAG motifs & Fan-out amplifies upstream failures; diamonds hide conditional
fragility. & Removing the reviewer corrector produces three downstream
cascades totaling $-0.087$; diamond fragility is $1.4\times$ when the shared
source fails. \\
Capacity & Recursive chain quality predicts output quality across depth and
catch rate. & Maximum theory--simulation gap is $0.002$ over 28 conditions. \\
Process entropy & Output quality should degrade approximately linearly in
$H(W)$ near a reference regime. & Fitted slopes are $\lambda\approx0.02$/bit;
higher $K/N$ increases intercepts and lowers $\lambda$. \\
Autonomy time & Drift-dominated first-passage time should scale as
$1/\mu$. & Log-log slope is $-0.99$ across two orders of magnitude in drift. \\
\bottomrule
\end{tabular}}
\end{table}

Across grid sizes ($6\times6$ to $20\times20$), catch rates
($0.30$--$0.90$), and skill distributions (uniform, Gaussian, bimodal), the
qualitative results persist. Quantitative values shift as expected: higher
catch rates increase masking, smaller grids increase trajectory noise, and
heterogeneous skills increase variance across the scope.

\section{Discussion}

The framework has three practical implications. First, delegated systems
must separate competence from delivered quality. A dashboard that records
only corrected output quality can look healthy while the producing agent is
weakening. The immediately implementable prescription is to log both
pre-correction and post-correction outcomes at every delegation boundary and
track $M^{*}=\sigma_{\mathrm{corr}}/\sigma_{\mathrm{raw}}$.

Second, topology is a governance variable. In chains, masking and quality
loss accumulate with depth. In fan-out structures, upstream correction is
amplified across children. In diamonds, average quality may hide conditional
fragility caused by a shared parent. The governance target should therefore
be chosen by sensitivity---for example
$\partial T^{*}_{\mathrm{auto}}/\partial c(v)$ or
$\partial C_{\mathrm{op}}/\partial c(v)$---rather than by local error rate
alone.

Third, autonomy has a finite buffer:
\[
    B_{\mathrm{eff}} = C_{\mathrm{op}}-p_{\min}-\lambda H(W).
\]
When this quantity is positive, the pipeline has room to operate without
continuous intervention; when it is near zero, small drift or added workflow
complexity can push the system below target. In the drift-dominated
approximation, the intervention cadence is
$T^{*}_{\mathrm{auto}}=B_{\mathrm{eff}}/\mu_{\mathrm{eff}}$.

\subsection*{Implementation}

A minimal operational version of the theory has four steps:
\begin{enumerate}\setlength\itemsep{2pt}
    \item Instrument each delegation boundary with raw and corrected outcome
    records.
    \item Estimate $\sigma_{\mathrm{raw}}$, $\sigma_{\mathrm{corr}}$,
    $M^{*}$, $H(W)$, and local drift over rolling windows.
    \item Check feasibility using $C_{\mathrm{op}}$ and
    $B_{\mathrm{eff}}$ before expanding autonomy.
    \item Allocate scarce human or model review to the node with the largest
    estimated gain in $C_{\mathrm{op}}$ or $T^{*}_{\mathrm{auto}}$.
\end{enumerate}
This is not a replacement for safety evaluation or deployment testing; it is a
calculus for deciding where such evaluation and oversight are most valuable.

\subsection*{Limitations and theoretical status}

The results have different levels of rigor. The water-filling allocation,
fixed points, masking index, capacity threshold, convergence time, and
recursive operational ceiling are derived under the stated Bernoulli
mean-field assumptions. The information-theoretic capacity theorem is exact
for the revealed-action stationary symbolwise governed channel. The
process-complexity law is a local Taylor approximation whose coefficient
$\lambda$ is measured or estimated. The autonomy-time result is a
drift-dominated first-passage scaling law; in the simulations it captures the
$1/\mu$ slope but overestimates absolute values by about 20\%.

The empirical validation is synthetic, not a production study. The agents do
not strategically adapt to the corrector; the corrector catch rate is fixed;
the outcomes are binary; the nodes are conditionally independent; and the
main experiments do not compare against strong baselines such as bandit
allocation, queue-aware control, active learning, or software-testing
heuristics. These are not small caveats. They define the current scope of the
claim: this paper proposes and internally validates a base theory of governed
delegation, not a universal theory of all agent coordination.

\subsection*{Conclusion}

Delegated AI autonomy is governed by a three-way tradeoff among competence,
governance capacity, and workflow complexity. The MSO makes that tradeoff
computable: allocate governed delegation by a Fisher-weighted variational
principle, separate raw competence from corrected quality, check whether the
system has enough capacity for the target quality, and schedule intervention
according to the remaining autonomy buffer. The central practical lesson is
simple: when an AI system delegates decisions to models, tools, or evaluators,
measure what the delegate can do before correction, measure what the system
delivers after correction, and do not confuse the two.

\section*{Acknowledgments and AI Assistance Disclosure}
\addcontentsline{toc}{section}{Acknowledgments and AI Assistance Disclosure}

This manuscript was written and revised with assistance from large
language models and AI-assisted coding tools, including Claude Code and
Codex, using models identified by the author as Claude Opus 4.7 and
GPT-5.5. These tools were used for drafting, editing, mathematical and
conceptual consistency checks, LaTeX revision, simulation-code support,
and preparation of reproducibility materials. They are tools, not
authors. The author reviewed, selected, and edited the resulting text,
derivations, code, figures, and references, and takes full responsibility
for the content of the paper. Any errors, omissions, incorrect claims, or
misinterpretations are the author's alone.


\end{document}